\documentclass{article}

\usepackage{arxiv}

\usepackage[utf8]{inputenc} % allow utf-8 input
\usepackage[T1]{fontenc}    % use 8-bit T1 fonts
\usepackage{hyperref}       % hyperlinks
\usepackage{url}            % simple URL typesetting
\usepackage{booktabs}       % professional-quality tables
\usepackage{amsfonts}       % blackboard math symbols
\usepackage{nicefrac}       % compact symbols for 1/2, etc.
\usepackage{microtype}      % microtypography
\usepackage{lipsum}
\usepackage{svg}
\usepackage{graphicx}
\graphicspath{ {./images/} }
\usepackage{amsmath} 
\usepackage{pdflscape}
\usepackage{xcolor}
\usepackage[normalem]{ulem}
\usepackage{comment}
\usepackage{longtable}
\usepackage{tabularx}
\usepackage{caption} % for more complex captioning
\usepackage{subcaption}
\usepackage{longtable}
\usepackage{booktabs}
\usepackage[utf8]{inputenc}
\usepackage{changepage}

\usepackage{array}
\usepackage{float}
\newlength{\extralength}
\newlength{\fulllength}
\newcolumntype{C}{>{\centering\arraybackslash}X}

\title{A Comparative Analysis of YOLOv5, YOLOv8, and YOLOv10 in Kitchen Safety}

\author{Athulya Sundaresan Geetha \textsuperscript{*} and Muhammad Hussain \\[1ex]
\begin{minipage}[t]{0.90\textwidth}
\centering
\scriptsize Department of Computer Science, Huddersfield University, Queensgate, Huddersfield HD1 3DH, UK; \\
\textsuperscript{*}Correspondence: U2282847@unimail.hud.ac.uk;
\end{minipage}}

\begin{document}

\maketitle
% Abstract (Do not insert blank lines, i.e. \\) 
\begin{abstract}Knife safety in the kitchen is essential for preventing accidents or injuries with an emphasis on proper handling, maintenance, and storage methods. This research presents a comparative analysis of three YOLO models, YOLOv5, YOLOv8, and YOLOv10, to detect the hazards involved in handling knife, concentrating mainly on ensuring fingers are curled while holding items to be cut and that hands should only be in contact with knife handle avoiding the blade. Precision, recall, F-score, and normalized confusion matrix are used to evaluate the performance of the models. The results indicate that YOLOv5 performed better than the other two models in identifying the hazard of ensuring hands only touch the blade, while YOLOv8 excelled in detecting the hazard of curled fingers while holding items. YOLOv5 and YOLOv8 performed almost identically in recognizing classes such as hand, knife, and vegetable, whereas YOLOv5, YOLOv8, and YOLOv10 accurately identified the cutting board. This paper provides insights into the advantages and shortcomings of these models in real-world settings. Moreover, by detailing the optimization of YOLO architectures for safe knife handling, this study promotes the development of increased accuracy and efficiency in safety surveillance systems.
\end{abstract}

% Keywords

\section{Introduction}
While working in the kitchen, one of the most common accidents is a knife mishap. It might happen to even the experienced cook if the knife is not handled properly with the utmost caution. Hazard detection when working with knife is important to prevent accidents and ensure adherence to safety guidelines. Safety knife handling includes the position of hands, fingers, and knife; in this paper, the main focus is on two hazards including curl fingers and hand touching blade in order to avoid cuts.

There are various challenges involved in the detection of hazards in the safe handling of knife. Variations in the light conditions in the kitchen can affect the way the knife appears in an image. The kitchen is often cluttered with electrical appliances, utensils, and ingredients, making the knife partially or fully obscured from the view. Training with a single knife makes the model accustomed to it and fails to recognize the different types of knives. Bad image qualities, involving low-resolution or blurry images, make it difficult for the models to find knife. The dynamic movement of the knife and hands can create challenges for real-time detection. Sometimes, tools in the kitchen might have a similar shape or color to a knife, which can be wrongly identified as knife. Effectively interpreting the hazard, for example, holding a knife wrongly, protruding the fingers that are placed on the item to be diced, or tapping on the knife, is quite difficult for the models.

Histogram of oriented gradients (HOG) was used to extract features from images both horizontally and vertically, while vertical histograms oriented gradients (VHOG) extracted features vertically \cite{RN1} and classified by support vector machine (SVM) and extreme learning machine (ELM). A machine learning technique, SVM, employed visualization of localized images and features during dynamic movements from spatiotemporal areas for classification but failed to recognize movements \cite{RN2, RN3, RN4}.

Because the two-stage approaches took a lot of time, depended on advanced architectures, and needed frequent manual supervision, the CNN models were introduced for detection and classification. Faster R-CNN alongside GoogleNet performed better in the detection of objects \cite{RN5}. In another study, three algorithms, MobileNet, MaskR-CNN, and PoseNet, were used to provide better performance, image segmentation, and hazard assessment, respectively. To improve object identification, more models like AlexNet \cite{RN6}, Google-Net \cite{RN7}, ResNet \cite{RN8}, VGG-Net \cite{RN9}, Region-based Convolutional Network method (R-CNN) \cite{RN10}, Fast-RCNN \cite{RN11}, and Faster-RCNN \cite{RN12} were designed \cite{RN13}.

Despite the improved performance, issues like efficient data handling, real-time execution, and better architectures continue. This is where YOLO (You Only Look Once) models come in handy to overcome the disadvantages that were faced by two-stage detection processes; built on DarkNet, YOLOv1 and Fast YOLO comprise 24 and 9 convolutional layers, respectively \cite{RN14}. YOLOv2 is an enhanced version, based on DarkNet-19, with standardization of batches, box references, depth class identification, fine-detailed features, and size-specific dimension clusters \cite{RN15}. Using DarkNet-53 architecture, YOLOv3 incorporates residual networks for the extraction of features, with binary cross-entropy \cite{RN16}. Along with YOLOv3’s predictive head, YOLOv4 uses CSPDarkNet53 as the backbone architecture and PANet with SPP as the neck network for better performance \cite{RN17}.

YOLOv5’s performance speed is obtained by adding an improved neck named CSP-PAN and a head named SPPF \cite{RN18}. In addition to performance speed, YOLv6 decreases the complexity of the computational requirements \cite{RN19}, whereas YOLOv7 with the integration of an auxiliary head and lead head increases accuracy in detecting objects \cite{RN20}. In YOLOv8, the framework supports object identification, image segmentation, and movement tracking by a modified CSPDarkNet53 backbone and PAN-FPN neck. Introducing Generalized Efficient Layer Aggregation Network and Programmable Gradient Information, both models, YOLOv9 and YOLOv10, improved the accuracy in the detection of objects; furthermore, in YOLOv10, non-maximum suppression was removed \cite{RN21, RN22}.

The aim of this study is to find a comparative analysis of the performance of YOLOv5, YOLOv8, and YOLOv10 in identifying hazards in handling knife safety in the kitchen environment. Performance metrics like accuracy, recall, mAP50, mAP50-95, F1 score, and confusion matrix are evaluated to find the best model among the three. This can be helpful in real-time to let the person know when the knife is incorrectly used majorly based on the safety guidelines to prevent mishaps.

The remainder of the paper is as follows: In Literature Review, the articles based on knife handling have been studied and elaborated especially on the basis of the YOLO models. The workflow and architectures of YOLO models, YOLOv5, YOLOv8, and YOLOv10, are detailed in the section titled Methodology. Exploratory performance matrices are discussed in Experimental Results and Discussion sections.

\section{Literature Review} 

Detection of hazards to ensure safety while handling knife involves recognizing potential dangers and taking measures to mitigate the risks. Support vector machines were used for offline classification after gathering local visual and movement features around spatiotemporal locations and their movement \cite{RN2, RN3}; however, they failed to recognize the activities involved in the preparation of food \cite{RN4}. Hence, using sensor readings from embedded devices within knives and spoons, a technique was introduced to identify actions, namely, dicing, peeling, mixing, and scooping, by extracting information on average, energy, variance, and entropy, including pitch and rotation for device positioning \cite{RN23}. While in-built Wii controllers were used in handles of utensils in the kitchen, another study not only added Wii controllers but also integrated these sensors alongside with RGBD camera, which was directed toward the workstation, to improve the performance of recognizing the activities \cite{RN24}. Images were pre-processed using linking size, shape, and position in the image to identify knife detectors by computer vision techniques, such as border identification, structure extraction, and segmentation. For detecting multiple features, Haar filters were utilized, where knives were determined correctly in 45\% of the cases and wrongly classified as knives in 85\% of the cases \cite{RN25}. However, there are limitations such as consuming more time, high-end architectures, and constant human interpretation.

Convolutional neural networks (CNN), such as AlexNet \cite{RN26}, Google-Net \cite{RN7}, ResNet \cite{RN8}, VGG-Net \cite{RN9}, Region-based Convolutional Network method (R-CNN) \cite{RN10}, Fast-RCNN \cite{RN11}, and Faster-RCNN \cite{RN12} were proposed and had a significant improvement in computer vision to enhance the detection of objects and classification of images \cite{RN13}. Knife and gun detection using the Faster R-CNN compared GoogleNet and SqueezeNet architectures, revealing GoogleNet achieved 46.68\% accuracy for knife detection and SqeezeNet performed better (85.44\%) for the detection of guns \cite{RN5}. The Faster R-CNN detected helmet wearing using Retinex for better outdoor image and K-means++ for small helmets, obtaining mAP of 94.3\% and a detection speed of 11.62 images per second \cite{RN40}. In a study by Noever and Noever (2020) \cite{RN27}, to determine the knife and threat detectors, three algorithms were used, where MobileNet classified with an accuracy of 95\%, MaskRCNN detected and segmented hands from knives, and PoseNet helped in positioning skeletal points in order to improve threat assessment and clarify intent.

The study proposed a detection and classification method for X-ray baggage security with 6 classes, whereby YOLO achieved 97.4\% for guns and ResNet reached 73.2\% for knife detection \cite{RN28}. In order to detect objects for visually impaired people in disorderly environments, YOLOv2 was used to detect, and shape-oriented methods were utilized to estimate three-dimensional models of objects in the kitchen and café room, providing safety instructions. With 90.45\% accuracy, the system processed at 0.86 frames per second \cite{RN29}. The kitchen interactions dataset was used to detect hands using YOLOv3, with VGG-16 achieving a higher Average Precision (AP) of 62.2\% when compared with MobileNet-Lite architecture; the limitations were blurred hands and noise \cite{RN30}. An improved lightweight YOLOv4 model was proposed to detect the management of real-time safety in on-site power work. To reduce model complexities and computational tasks, it employed a state-of-the-art backbone network with depthwise separable convolutions and mobile-inverted bottleneck structures; to extract multiple features and extract small objects, an enhanced bidirectional fusion network was introduced. In addition, training and detection accuracy was enhanced by improving the confidence loss. It achieved a significant reduction in a number of parameters (93.11\%), increased detection speed (22\%), and obtained 84.82\% accuracy in facilitating real-time safety \cite{RN31}.

Utilizing YOLOv5 and Kernelized Correlation Filter, SafeCOOK, a real-time, low-cost system, tracked kitchen appliances, enhancing safety by identifying dangerous scenarios during cooking. YOLOv5 alone lost overlapping objects; YOLOv5 with Single Object Tracking and with Multiple Object Tracking tracked only one type and multiple objects, respectively, but with errors; and YOLOv5 with Multiple Object Tracking and dropout tracking tacked consistently without many errors \cite{RN32}. To accurately detect whether the kitchen staff wore masks correctly, the YOLOv5 model performed well with mAP50 of 97.6\%, with full accuracy in 89 epochs and an inference speed of 10 frames per second \cite{RN33}. The authors proposed an augmented reality assistance for cooking utilizing in-built sensors through the movements of hands. Although YOLOv5 model achieved lower accuracy than models, ResNet and ResNeXt, it classified multiple categories at once. Future works will focus on enhancing the machine by monitoring real-time cooking, improving the identification of hand gestures, developing models to detect failures, and integrating models to assist in recovery from mistakes \cite{RN34}. A real-time object detection based on YOLOv5 was developed to control heater power or alert users by detecting scenarios like tracking boiling liquids, utensils arrangement, and empty stovetops. YOLOv5 attained over 98\% precision and recall. In future work, the emphasis will be on the integration of microcontrollers in electronics and the expansion of the object categories \cite{RN35}.

Moreover, to achieve high speed and better accuracy in identifying items related to fire, the authors presented the YOLOv6 model to detect fire in smart cities. YOLOv6 demonstrated high performance (98\%), recall (96\%), and precision (83\%), identifying fire-related objects within 0.66 seconds. In the coming years, the primary goal will be to improve performance in challenging scenarios and develop a model, 3D CNN/U-Net, to use in different sectors \cite{RN36}. The detection of weapons like knives and guns for the safety of people was performed by YOLOv7 and YOLOv8 architectures, providing accurate detection of weapons through bounding box annotations. YOLOv7-e6 outperformed other models by achieving a high mAP of 90.3\% at 0.5 IoU threshold. Future works will explore YOLOv8 for enhancement of determining the objects in real time in different sectors \cite{RN37}. A study utilized YOLOv8, first testing the pre-trained model on images for knife detection and training a custom model to recognize knives \cite{RN38}. In another study, YOLOv8 was used to detect hygiene violations, such as improper dress and pests, with 89\% accuracy, ensuring the food safety and health of the customers \cite{RN39}.

To conclude, this literature review provides the application of YOLO models majorly in the kitchen, ranging from efficient detection of knives, movement of hands, hygiene, to safety measures. From all the YOLO models, it is noted that there is an enhancement not only in performance and speed but also in flexibility, for example, blurred images due to dynamic movement and objects that are small or hidden are handled better with each new version. Although different scenarios were used in the kitchen setting, there is no study on the safety of knife handling. This paper deals with the 2 main hazards, curl fingers and hand touching blade, and compares the performance of the YOLO models, YOLOv5, YOLOv8, and YOLOv10, while using knives.

\section{Methodology}
\subsection{Dataset}
\begin{figure}[H]
\begin{adjustwidth}{-\extralength}{0cm}
\centering
\includegraphics[width=15cm]{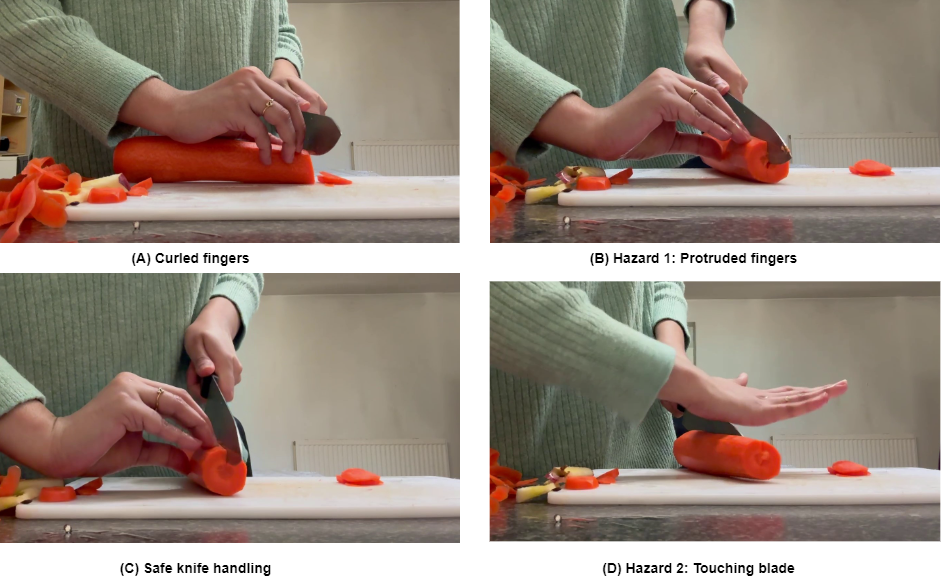}
\end{adjustwidth}
\caption{Sample image of the kitchen dataset.}
\label{Figure:1}
\end{figure}

The dataset was derived from a high-resolution video (1920 × 1080 pixels) recorded using an Apple iPhone15 Pro. Then, the video was processed and converted into 6004 individual frames using a Python script. The extracted frames were then manually annotated utilizing Label Studio with the following 6 classes: cutting board, hands, vegetable, knife, hazard 1: curl finger, and hazard 2: hand touching blade. The safe use of knives in the kitchen was examined, where hazard 1 was designated as curling finger to prevent accidental cuts, and hazard 2 was identified as the hand coming in contact with the blade to avoid serious hand injuries. Sample images of the collected dataset are shown in Figures \ref{Figure:1}A-D, showing the supporting hand's finger positions, curled (Figure \ref{Figure:1}A) and protruded (Figure \ref{Figure:1}B), as well as safe knife handling (Figure \ref{Figure:1}C) and hand contact with the blade (Figure \ref{Figure:1}D).

\subsection{Data Augmentation}
Data augmentation of images is a crucial technique in the preprocessing stage of image analysis in machine learning and computer vision to improve the robustness and performance of the model. Due to the limited size of the dataset, image augmentation was employed to generate numerous additional images. Applying transformations introduced variability in the training data, preparing models for new, unseen data. With different training images, the model would not fall into the category of memorizing particular details, thereby reducing overfitting. In addition, augmentation addresses class imbalance by adding more images to the underrepresented classes, improving the accuracy of the overall classification and reducing the bias toward the majority classes. Overall, data augmentation not only increases the dataset diversity but also promotes the development of machine learning models to tackle real-world difficulties effectively.

Detecting hazards in the kitchen while handling knives poses challenges namely fluctuating light conditions, dynamic hand gestures, blurred images, obstructed views, and diverse angles. These challenges have been addressed by incorporating a wider range of images to the training dataset. By augmenting the images with variations in lighting conditions and potential occlusions, the model is trained to adapt to different situations. This helps the model to perform better, enhancing its ability to detect hazards. For example, variations in the light create different kitchen environments, thereby preparing the model to work under different lighting settings. Adding occlusions mimics real-world challenges, making the model detect hazards even when the objects are partially obstructed. Augmentation ensures that the model is capable of handling the complexities of kitchen environments, contributing to safer knife-handling practices.

\subsubsection{Random Crop}
Cropping is utilized to improve the model’s ability to detect hazards while safe knife handling, ensuring the detection of partial visibility of the objects, uniform scaling, or varied distances from the camera. The cropping process adjusts images from 0\% minimum zoom, retaining original dimensions, to 20\% maximum zoom, percentage of image dimensions to crop from each sides. Hence, the sizes of the images range from 512 × 512 to 640 × 640 pixels, adapting real-world variations in the knife usage. Equation \ref{eq:1} is utilized to crop 20\% from each side of the images.

\begin{equation}
    D^{\prime}=D[ top +p \times height, bottom -p \times height, left +p \times width, right -p
    \label{eq:1}
\end{equation}

\ \(D^{\prime}\) is the resulting image; \(D\) is the original image; and \(p\) represents the cropping percentage of 20\%.

\subsubsection{Random Rotation}
Rotation in this dataset involves varying the position of the knife within a specified range (i.e., ±15 degrees) (Equation \ref{eq:2}). This technique enhances the robustness of the model to detect knife from different angles, enhancing its ability to detect and handle knife-related hazards effectively in kitchen scenarios.

\begin{equation}
    D^\prime\ =\ rotate(D,\theta)
    \label{eq:2}
\end{equation}

Equation \ref{eq:3} displays the 2-dimensional rotational transformation of an image.

\begin{equation}
    R\left(\theta\right)=\left[\begin{matrix}cos(\theta)&-sin(\theta)\\sin(\theta)&cos(\theta)\\\end{matrix}\right]
    \label{eq:3}
\end{equation}

Rotation matrix is represented by \ \( R(\theta) \).

\subsubsection{Random Shear}
This technique involves slanting the image along the x or y axis, creating a tilt that distorts objects in the images. For safe knife usage, random shearing provides conditions where the knife or hands appear slanted (±10 degrees) due to different angles, resulting in geometric distortion (Equation \ref{eq:4}).

\begin{equation}
    D^\prime\ =\ D\cdot S
    \label{eq:4}
\end{equation}

Shear matrix is denoted by \textit{S}.

\subsubsection{Random Grayscale}
Grayscaling, applied to 15\% of colored images, improves performance of the model by taking into consideration the shapes and textures rather than colors. This technique is used to provide varied lighting conditions, low-light or monochromatic images, to improve the model’s ability to detect hazards in different settings (Equation \ref{eq:5}).

\begin{equation}
    D' = 
    \begin{cases} 
        0.299 \times R + 0.587 \times G + 0.114 \times B, & \text{with probability } p \\
        D, & \text{with probability } (1-p)
    \end{cases}
    \label{eq:5}
\end{equation}

\(R\) denotes red, \(G\) means green, and \(B\) indicates blue; \(p\) represents the transformation probability.

\subsubsection{Hue}
Hue has been used to get different color variations in the images. A range of -15\% to +15\% denotes the variations in colors that span from a baseline value of -15\% to a maximum value of 15\% (Equation \ref{eq:6}).

\begin{equation}
    D^\prime=D+H
    \label{eq:6}
\end{equation}

In this augmentation formula, \(H\) is the hue shift factor ranging from -15\% to +15\%.

\subsubsection{Saturation}
Saturation from -25\% to +25\% exposes images to different color intensities, providing different lighting settings and color variations (Equation \ref{eq:7}).

\begin{equation}
    D^\prime\ =\text{adjust\_saturation}(D,\alpha)
    \label{eq:7}
\end{equation}

$\alpha$ is the saturation adjustment factor.

\subsubsection{Brightness}
Here, brightness has been adjusted in the range of -15\% to +15\% in order to expose the images to different lighting conditions (Equation \ref{eq:8}). This helps the model to adapt to various brightness levels to detect hazards correctly in different environments.

\begin{equation}
    D^\prime\ =\ clip(D\ +\ \beta\ \times255,\ 0,255)
    \label{eq:8}
\end{equation}

\ \(\beta\) is an adjusting factor for brightness.

\subsubsection{Exposure}
Exposure means increasing or decreasing the amount of light captured in an image. By adjusting the exposure from -10\% to +10\%, it decreases the brightness by up to -10\% and increases the brightness to +10\% from the original image (Equation \ref{eq:9}).

\begin{equation}
    D^\prime=D\times\left(1+\frac{A}{100}\right),
    \label{eq:9}
\end{equation}

where \(A\) is the exposure adjustment factor (-10\% to +10\%).

\subsubsection{Blur}
Blurring applies a blur effect to images to enhance the performance of the model to generate real-world settings where images might not be always sharp and contrast. So, the model learns to detect hazards even in out-of-focus objects, knife or hands, to identify risks in various scenarios. It also decreases the noise as well as increases the number of features to be extracted. The standard deviation of the Gaussian blur kernel is illustrated by $\sigma$ (Equation \ref{eq:10}).

\begin{equation}
    D^\prime\ =\ Blur(D,\sigma),
    \label{eq:10}
\end{equation}

\subsubsection{Random Noise}
Adding noise to the images helps the model to learn important features even when there are distortions (Equation \ref{eq:11}). In this approach, 0.1\% of the pixels in the image are affected by noise.

\begin{equation}
    D^\prime\ =\ D+N
    \label{eq:11}
\end{equation}

\subsubsection{HSV}
HSV stands for hue, saturation, and value, in which hue denotes type of the color, saturation indicates color intensity, and value means the brightness of the image. Value is set to 0.4 to reduce the brightness to 40\%. Hue is set to 0.015 to create color variations. Saturation sets at 0.7 to increase the saturation to 70\% to enhance the intensity of the color. $\sigma$ is the brightness adjusted on the pixel (Equation \ref{eq:12}).

\begin{equation}
    D^\prime\ =\ \alpha D
    \label{eq:12}
\end{equation}

\subsubsection{Translate}
Translate shifts the image along the vertical and horizontal axes. With translate set to 0.1, the image can be repositioned at width and height by 10\%, by horizontally and vertically. The translation factors are represented by $t_x$ and $t_y$ (Equation \ref{eq:13}).

\begin{equation}
    D^\prime\left(x+t_x\times w i d t h,\ y+t_y\ \times h e i g h t\right)=D(x,\ y)
    \label{eq:13}
\end{equation}

\subsubsection{Mosaic}
Mosaic augmentation combines four different images into a single mosaic image. When mosaic=1.0 is set, it means applying mosaic to 100\% of the training images to enhance the variety and complexity of the training data. Mosaic image coordinates are represented by $(x^\prime)$ and $(y^\prime)$; the original image corresponding coordinates are denoted by \( x_i \) and \( y_i \); the intensity of the image is set as \( I_i \) (Equation \ref{eq:14}).

\begin{equation}
    D^\prime\left(x^\prime,\ y^\prime\right)=\ D_i(x_i,\ y_i)
    \label{eq:14}
\end{equation}

\subsubsection{Random Erasing}
By using erasing method, parts of the image can be removed or hidden. Here, the value of 0.4 signifies that 40\% of the image is erased. After erasing process, the intensity of the pixel is indicated by \( D^\prime(x,\ y) \) (Equation \ref{eq:15}).

\begin{equation}
   D^\prime(x,\ y)=\left\{\begin{matrix}0,&if\ \left(x,\ y\right)is\ within\ the\ erased\ region\\D\left(x,\ y\right),&otherwise\\\end{matrix}\right.
    \label{eq:15}
\end{equation}

Data augmentation is crucial in detecting hazards when using a knife, as it helps create large datasets and avoid overfitting by creating different scenarios for the model to learn from.

\subsection{YOLOv5 Architecture}

The architecture of the YOLOv5 model consisted of a backbone, neck, and head. In the backbone, YOLOv5 utilizes a ResNet-based cross-stage partial (CSP) net, increasing efficiency through cross-stage partial connections and implementing multiple spatial pyramid pooling (SPP) blocks to extract multiple features, reducing the load of the computational tasks. A path aggregation network (PAN) module and upscaling layers are integrated into the neck to boost the resolution of the feature map and optimize the exchange of the information flow between the layers. Comprising 3 convolutional layers, the head of YOLOv5 is employed for the prediction of bounding boxes, class predictions, and confidence scores \cite{RN41}.

Anchor-based predictions are used to link the bounding box to the predefined anchor boxes with particular dimensions and forms. The loss function in YOLOv5 has Binary Cross-Entropy for class and objectness losses, and Complete Intersection over Union (CIoU) for location loss, with objectness loss calculation method dependent on the size of the prediction layer \cite{RN13, RN41}. YOLOv5 architecture including layers with a number of filters, size specifying the filter size, and repeat showing how many times the layer is used is given in Table \ref{tab:yolov5_architecture}.

\begin{table}[H]
\caption{YOLOv5 Architecture.\label{tab:yolov5_architecture}}
\begin{adjustwidth}{-\extralength}{0cm}
    \begin{tabularx}{\fulllength}{CCCCCC}
        \toprule
        \textbf{Layer} & \textbf{Activation} & \textbf{Filters} & \textbf{Size} & \textbf{Repeat} & \textbf{Output Size} \\
        \toprule
        Image          & -          & -       & -          & -      & 640 × 640 \\
        Conv0          & ReLU       & 16      & 3 × 3 / 2  & 1      & 320 × 320 \\
        Conv1          & ReLU       & 32      & 3 × 3 / 2  & 1      & 160 × 160 \\
        Conv2          & ReLU       & 64      & 3 × 3 / 2  & 1      & 80 × 80  \\
        Conv3          & ReLU       & 128     & 3 × 3 / 2  & 1      & 40 × 40  \\
        Conv4          & ReLU       & 256     & 3 × 3 / 2  & 1      & 20 × 20  \\
        C3           & ReLU       & 128     & 1 × 1      & 1      & 20 × 20  \\
        Conv5          & ReLU       & 128     & 3 × 3      & 1      & 20 × 20  \\
        C3           & ReLU       & 256     & 1 × 1      & 1      & 20 × 20  \\
        Conv6          & ReLU       & 256     & 3 × 3      & 1      & 20 × 20  \\
        SPP            & -          & -       & -          & -      & 20 × 20  \\
        C3           & ReLU       & 512     & 1 × 1      & 1      & 20 × 20  \\
        Upsample       & -          & -       & -          & 1      & 40 × 40  \\
        C3           & ReLU       & 256     & 1 × 1      & 1      & 40 × 40  \\
        Upsample       & -          & -       & -          & 1      & 80 × 80  \\
        C3           & ReLU       & 128     & 1 × 1      & 1      & 80 × 80  \\
        Conv7          & ReLU       & 128     & 3 × 3      & 1      & 80 × 80  \\
        C3           & ReLU       & 64      & 1 × 1      & 1      & 80 × 80  \\
        Conv8          & ReLU       & 64      & 3 × 3      & 1      & 80 × 80  \\
        C3           & ReLU       & 32      & 1 × 1      & 1      & 80 × 80  \\
        \bottomrule
    \end{tabularx}
\end{adjustwidth}
\end{table}

In detail, the model uses a series of convolutional layers, with ReLu activation to reduce the dimensions from 640 × 640 to 320 × 320 to 20 × 20 pixels. Following this, the C3 layer consists of a combination of 1 × 1 and 3 × 3 convolutions repeated many times, without changing the size of the output. Spatial pyramid pooling (SPP) layer pools features from different scales while maintaining a dimension of 20 × 20 pixels. Upscaling increases the feature maps to a higher resolution.

When compared to the previous versions, YOLOv5 aimed at increasing performance and accuracy, introducing enhancement in extracting multiple features, aggregating, and predicting based on the anchors. To make it to adapt to iOS devices, ONNX and CoreML frameworks have been used rather than PyTorch. Achieving a 50.7\% average precision on MS COCO dataset, YOLOv5x processed 640-pixel images at 200 frames per second, with a higher average precision of 55.8\% at a 1534-pixel image size.

\subsection{YOLOv8 Architecture}
Based on YOLOv5, YOLOv8 incorporates architectural improvements for superior detection accuracy. YOLOv8 uses the C2f module to transition to an anchorless architecture, which simplifies the prediction of bounding boxes and decreases the reliance on non-maximum suppression by removing traditional anchor boxes that misrepresent various patterns in the dataset \cite{RN41}. The architectural table not only provides the details of the configuration of each layer, incorporating the C2f module and Spatial Pyramid Pooling Fractional (SPPF) to increase the accuracy of the location and feature representation, but also emphasizes the type and size of the filters, and dimensions of the output.

The model processes the input image through a sequence of convolutional layers, Conv0, Conv2, C2f2, and C2f4, using ReLU activation to reduce the pixel size from 320 × 320 to 80 × 80. Furthermore, the size is reduced to 20 × 20 in Conv5 to Conv7. Upscaling layers increase the size from 20 × 20 to 80 × 80 pixels. Finally, the remaining convolutional layers help in detecting the object (Table \ref{tab:yolov8_architecture}).

\begin{table}[H]
\caption{YOLOv8 Architecture.\label{tab:yolov8_architecture}}
\begin{adjustwidth}{-\extralength}{0cm}
    \begin{tabularx}{\fulllength}{CCCCCC}
        \toprule
        \textbf{Layer} & \textbf{Activation} & \textbf{Filters} & \textbf{Size} & \textbf{Repeat} & \textbf{Output Size} \\
        \toprule
        Image          & -          & -       & -          & -      & 640 × 640 \\
        Conv0          & ReLU       & 32      & 3 × 3 / 2  & 1      & 320 × 320 \\
        Conv1          & ReLU       & 64      & 3 × 3 / 2  & 1      & 160 × 160 \\
        C2f2           & ReLU       & 64      & 3 × 3 / 2  & 2      & 160 × 160 \\
        Conv3          & ReLU       & 128     & 3 × 3 / 2  & 1      & 80 × 80  \\
        C2f4           & ReLU       & 128     & 3 × 3 / 2  & 4      & 80 × 80  \\
        Conv5          & ReLU       & 256     & 3 × 3 / 2  & 1      & 40 × 40  \\
        C2f6           & ReLU       & 256     & 3 × 3 / 2  & 6      & 40 × 40  \\
        Conv7          & ReLU       & 512     & 3 × 3 / 2  & 1      & 20 × 20  \\
        C2f8           & ReLU       & 512     & 3 × 3 / 2  & 8      & 20 × 20  \\
        SPPF           & ReLU       & 512     & -          & 1      & 20 × 20  \\
        Upsample10     & -          & -       & -          & 1      & 40 × 40  \\
        Concat11       & ReLU       & 512     & -          & 1      & 40 × 40  \\
        C2f12          & ReLU       & 256     & 3 × 3      & 12     & 40 × 40  \\
        Upsample13     & -          & -       & -          & 1      & 80 × 80  \\
        Concat14       & ReLU       & 256     & -          & 1      & 80 × 80  \\
        C2f15          & ReLU       & 128     & 3 × 3      & 15     & 80 × 80  \\
        Conv16         & ReLU       & 128     & 3 × 3      & 1      & 80 × 80  \\
        Concat17       & ReLU       & 128     & -          & 1      & 80 × 80  \\
        C2f18          & ReLU       & 64      & 3 × 3      & 18     & 80 × 80  \\
        Conv19         & ReLU       & 64      & 3 × 3      & 1      & 80 × 80  \\
        Concat20       & ReLU       & 64      & -          & 1      & 80 × 80  \\
        C2f21          & ReLU       & 32      & 3 × 3      & 21     & 80 × 80  \\
        Detect22       & -          & -       & -          & -      & 80 × 80  \\
        \bottomrule
    \end{tabularx}
\end{adjustwidth}
\end{table}

YOLOv8 employs sigmoid functions, evaluating the presence of an object in a bounding box, and softmax functions, categorizing the object into its respective class. Unlike previous YOLO versions, by incorporating Complete Intersection over Union and distribution focal loss functions, the model improvements enhance the accuracy in predicting bounding box and address the detection of smaller models. Binary cross-entropy is employed to calculate classification loss, thereby enhancing the classification of multiple classes.

YOLOv8x shows better performance on the dataset, MS COCO test-dev 2007, obtaining an average precision of 53.9\% for 640-pixel images operating at a speed of 280 frames per second when TensorRT is used.

\subsection{YOLOv10 Architecture}
YOLOv10 sets a new benchmark by increasing inference speed, eliminating non-maximum suppression, and introducing a dual-label assignment system \cite{RN41}. It balances supervision and rapid inference through the one-to-many and one-to-one branches, respectively, and improving the effectiveness of training and inference tasks by applying a standardized matching technique to maintain uniformity in those 2 branches.

The YOLOv10 model analyses the image using a set of convolutional layers, Conv0, Conv1, C2f2, and C2f4, using ReLU activation, decreasing the output size from 320 × 320 to 80 × 80 pixels. Further layers, namely SCDown5, C2f6, and C2f8, reduce the pixel size to 20 × 20. The Spatial Pyramid Pooling and Pyramid Squeeze Attention layers refine the features. Then, upsampling layers increase the size to 80 × 80. The detailed architecture of YOLOv10 with layers, filters, and outputs is displayed in Table \ref{tab:yolov10_architecture}.

\begin{table}[H]
\caption{YOLOv10 Architecture.\label{tab:yolov10_architecture}}
\begin{adjustwidth}{-\extralength}{0cm}
    \begin{tabularx}{\fulllength}{CCCCCC}
        \toprule
        \textbf{Layer} & \textbf{Activation} & \textbf{Filters} & \textbf{Size} & \textbf{Repeat} & \textbf{Output Size} \\
        \toprule
        Image          & -          & -       & -          & -      & 640 × 640 \\
        Conv0          & ReLU       & 32      & 3 × 3 / 2  & 1      & 320 × 320 \\
        Conv1          & ReLU       & 64      & 3 × 3 / 2  & 1      & 160 × 160 \\
        C2f2           & ReLU       & 64      & 3 × 3 / 2  & 2      & 160 × 160 \\
        Conv3          & ReLU       & 128     & 3 × 3 / 2  & 1      & 80 × 80  \\
        C2f4           & ReLU       & 128     & 3 × 3 / 2  & 4      & 80 × 80  \\
        SCDown5        & ReLU       & 128     & 3 × 3 / 2  & 1      & 80 × 80  \\
        C2f6           & ReLU       & 256     & 3 × 3 / 2  & 6      & 40 × 40  \\
        SCDown7        & ReLU       & 256     & 3 × 3 / 2  & 1      & 40 × 40  \\
        C2f8           & ReLU       & 512     & 3 × 3 / 2  & 8      & 20 × 20  \\
        SPPF9          & ReLU       & 512     & -          & 1      & 20 × 20  \\
        PSA10          & ReLU       & 512     & -          & 1      & 20 × 20  \\
        Upsample11     & -          & -       & -          & 1      & 40 × 40  \\
        Concat12       & ReLU       & 512     & -          & 1      & 40 × 40  \\
        C2f13          & ReLU       & 256     & 3 × 3      & 13     & 40 × 40  \\
        Upsample14     & -          & -       & -          & 1      & 80 × 80  \\
        Concat15       & ReLU       & 256     & -          & 1      & 80 × 80  \\
        C2f16          & ReLU       & 128     & 3 × 3      & 16     & 80 × 80  \\
        Conv17         & ReLU       & 128     & 3 × 3      & 1      & 80 × 80  \\
        Concat18       & ReLU       & 128     & -          & 1      & 80 × 80  \\
        C2f19          & ReLU       & 64      & 3 × 3      & 19     & 80 × 80  \\
        SCDown20       & ReLU       & 64      & 3 × 3      & 1      & 80 × 80  \\
        Concat21       & ReLU       & 64      & -          & 1      & 80 × 80  \\
        C2fCIB22       & ReLU       & 32      & 3 × 3      & CIB    & 80 × 80  \\
        v10Detect23    & -          & -       & -          & -      & 80 × 80  \\
        \bottomrule
    \end{tabularx}
\end{adjustwidth}
\end{table}

With the layers, that is, partial self-attention (PSA) and spatially constrained (SC), this architecture improves precision and performance, and includes depthwise, decreasing resolutions, and pointwise, increasing the dimensions of the channel, convolutions and a dual-label assignment for training and inferences optimization, thus allowing YOLOv10 model to handle the detection of occluded or small objects with better speed and precision.

YOLOv10 model is structured in a way to enhance efficiency without compromising performance by using a lightweight classification head to tackle the bottleneck issue associated with regression tasks. About 70\% accuracy was obtained while integrating depthwise convolution and an additional 30\% while the same is added into a compact inverted block, further combined with rank-guided block design.

\section{Experimental Results}
Performance metrics of YOLOv5, YOLOv8, and YOLOv10 during both the training and validation phases are discussed in this section. The training and validation of all three models were conducted in a PyTorch environment with NVIDIA GPUs, a system with high performance. In order to improve the accuracy and performance of the models, hyperparameters were tuned over the course of 40 epochs. The training utilized the AdamW optimizer with a 0.001 learning rate and 0.9 momentum. Additionally, group parameters were fine-tuned in a way where no decay was seen in some weights, whereas biases received L2 regularization with a decay rate of 0.0005, preventing overfitting while retaining generalizability.

\begin{figure}[H]
\begin{adjustwidth}{-\extralength}{0cm}
\centering
\includegraphics[height=8cm]{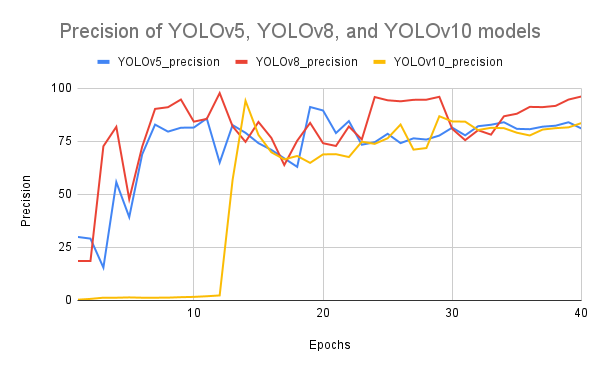}
\end{adjustwidth}
\caption{Comparison of precision values across three models.}
\label{Figure:2}
\end{figure}

From Figure \ref{Figure:2}, when the training set was trained to 40 epochs, it can be noted that the training precision curves for all 3 models converged after 13 epochs. YOLOv5 improved significantly stabilizing between 60\% and 80\%. YOLOv8 showed the highest accuracy, increasing and stagnant around epoch 7. Although YOLOv10 was not performing well in the beginning, after 13 epochs, it increased around 0.7 and fluctuated around 60\% and 80\%, showing less stability. Finally, it is noted that the YOLOv8 had the highest training accuracy throughout the 40 epochs when compared to other models.

\begin{figure}[H]
\begin{adjustwidth}{-\extralength}{0cm}
\centering
\includegraphics[height=8cm]{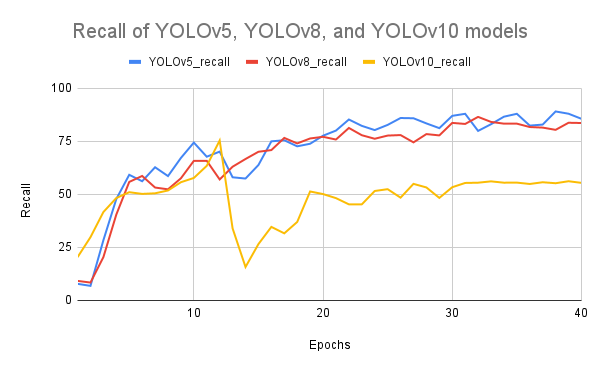}
\end{adjustwidth}
\caption{Comparison of recall values across the three models.}
\label{Figure:3}
\end{figure}

As shown in Figure \ref{Figure:3}, the recall graph compared YOLOv5, YOLOv8, and YOLOv10 over 40 epochs. Just like precision, YOLOv5 improved steadily and stabilized between 60\% and 80\% with some fluctuations. YOLOv8 showed the best performance, increasing around 80\%. YOLOv10 illustrated variability, increased at 7, and then dropped consistently, indicating a less stable performance of the model.

\begin{figure}[H]
\begin{adjustwidth}{-\extralength}{0cm}
\centering
\includegraphics[height=10cm]{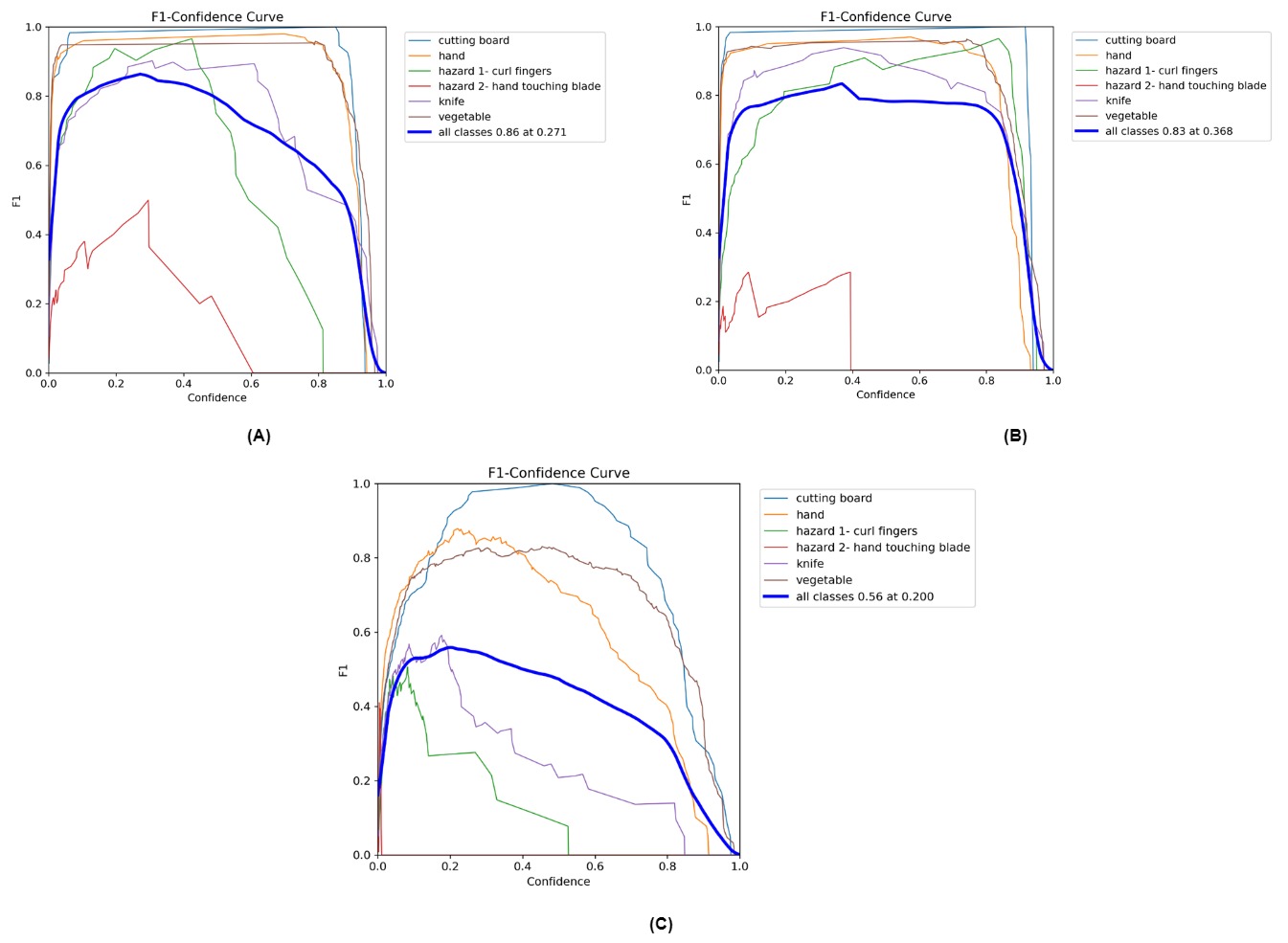}
\end{adjustwidth}
\caption{Comparison of F1 versus confidence across the three models.}
\label{Figure:4}
\end{figure}

The model’s F1-confidence curves displayed the balance between precision and recall at different confidence levels, addressing safe knife handling with classes such as cutting board, hand, hazard 1: curl fingers, hazard 2: hand touching knife, knife, and vegetable. The F1-confidence curves for all models with different classes are illustrated in Figures \ref{Figure:4}A, \ref{Figure:4}B, \ref{Figure:4}C.

For the YOLOv5 model, the classes, namely cutting board, hand, knife, and vegetable, had high F1 scores with various confidence levels. On the contrary, the hazard classes (curl fingers and hand touching blade) had lower F1 scores, especially hazard 2 as the performance dropped with increasing confidence levels.

The combined F1 score of all classes peaked at 0.86 at a confidence level of 0.271 for YOLOv5, thereby providing a balanced trade-off between accuracy and recall. It means that at this point, there would be minimal false positives or negatives with better accuracy. However, YOLOv8 had a slightly lower peak F1 score of 0.83 at a confidence interval of 0.368 and YOLOv10 had the lowest peak F1 score of 0.56 at 0.200. These results indicated that YOLOv5 outperformed the other two models.

In YOLOv5, after 0.4 F1 rate, the confidence of all classes had increased till 0.3 and then decreased gradually, whereas the confidence of all classes fluctuated at 0.4 and then started to gradually decrease in YOLOv8. The confidence of all classes in YOLOv10 had increased from 0.1 to 0.2 and then decreased afterward.

From all the graphs, it is evident that all the classes increased as the confidence level increased to a certain point and then decreased after certain confidence levels. In YOLOv5 and YOLOv8, class hazard 2 had the lowest F1 rate with less confidence level, while in YOLOv10, hazard 1 class had the lowest F1 and confidence level rate. In all three models, the cutting board class had the highest F1 rate.

\begin{figure}[H]
\begin{adjustwidth}{-\extralength}{0cm}
\centering
\includegraphics[height=6cm]{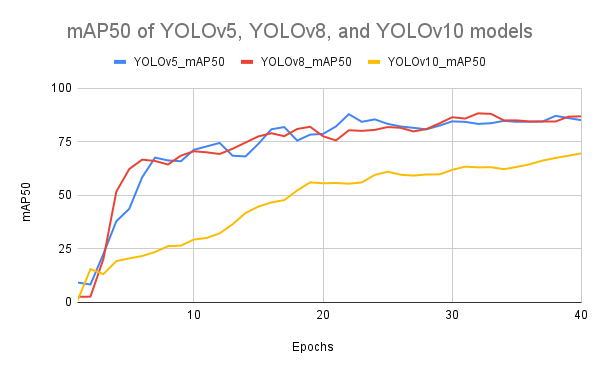}
\end{adjustwidth}
\caption{Comparison of mAP50 values across three models.}
\label{Figure:5}
\end{figure}

\begin{figure}[H]
\begin{adjustwidth}{-\extralength}{0cm}
\centering
\includegraphics[height=6cm]{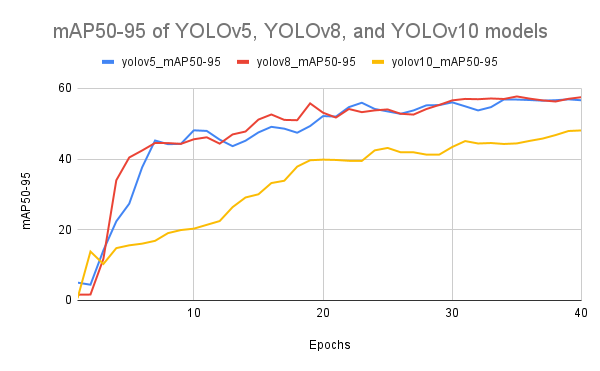}
\end{adjustwidth}
\caption{Comparison of mAP50-95 values across three models.}
\label{Figure:6}
\end{figure}

Initially, YOLOv8 reached 50\% for mAP50 (Figure \ref{Figure:5}) and 30\% for mAP50-95 (Figure \ref{Figure:6}) by epoch 5. YOLOv5 followed YOLOv8 till epoch 20, fluctuating between 70\% and 80\% for mAP50 and between 50\% and 60\% for mAP50-95. YOLOv10 improved at a steady pace; by epoch 40, it achieved 60\% for mAP50 and 45\% for mAP50-95. Overall, YOLOv8 was the effective model for overall mAP performance.

Classification accuracy for all the classes in the three models, YOLOv5, YOLOv8, and YOLOv10, can be provided by the normalized confusion matrix. In all three models, the cutting board class was classified correctly with the highest accuracy of 100\%, indicating that the models can effectively identify the category without error.

YOLOv5 and YOLOv8 classified hand correctly with a high accuracy of 0.96, whereas the classification score of the YOLOv10 model was 0.88, denoting some challenges in correctly identifying this class. 

The hazard 1 class demonstrated a difference in the performance of the models. YOLOv8 performed well with an accuracy of 1.00. In contrast, YOLOv5 and YOLOv10 showed significant misclassification with accuracies of 0.50 and 0.16, respectively. These indicated that the YOLOv8 model had a better ability to recognize hazard 1

Unlike other categories, all the models performed poorly for the category hazard 2 with accuracies of 0.50, 0.33, and 0.27, respectively, for YOLOv5, YOLOv8, and YOLOv10. This means that it was difficult for all models to recognize hazard 2.

\begin{figure}[H]
\begin{adjustwidth}{-\extralength}{0cm}
\centering
\includegraphics[height=10cm]{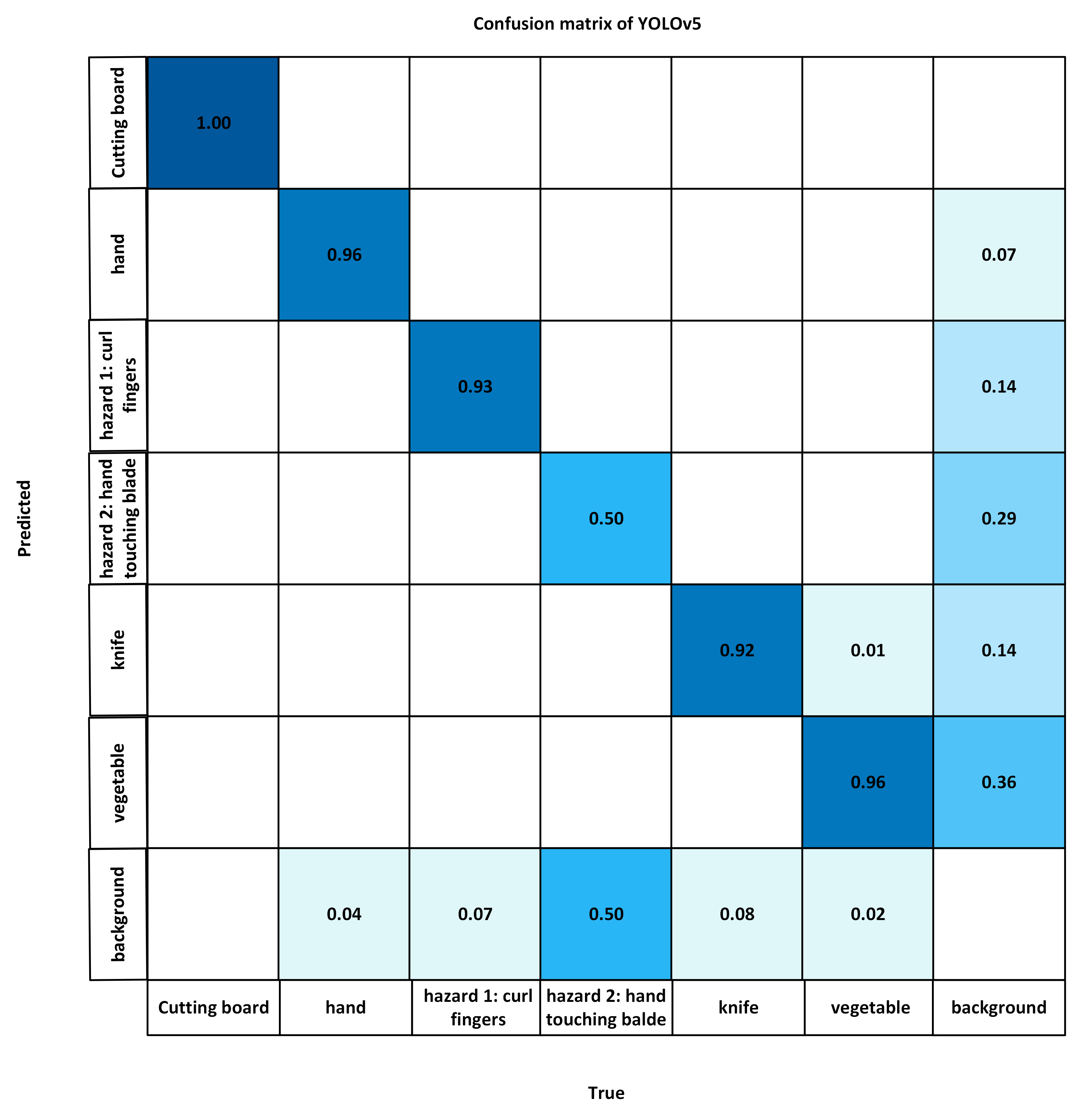}
\end{adjustwidth}
\caption{YOLOv5 normalized confusion matrix.}
\label{Figure:7}
\end{figure}

In the knife category, YOLOv5 (Figure \ref{Figure:7}) and YOLOv8 performed similarly, with an accuracy of 0.92; on the other hand, YOLOv10 was slightly lower at 0.89. Vegetable showed high accuracy in YOLOv5 and YOLOv8, with 0.96 and 0.95, respectively, while YOLOv10 dropped to 0.89.

\begin{figure}[H]
\begin{adjustwidth}{-\extralength}{0cm}
\centering
\includegraphics[height=10cm]{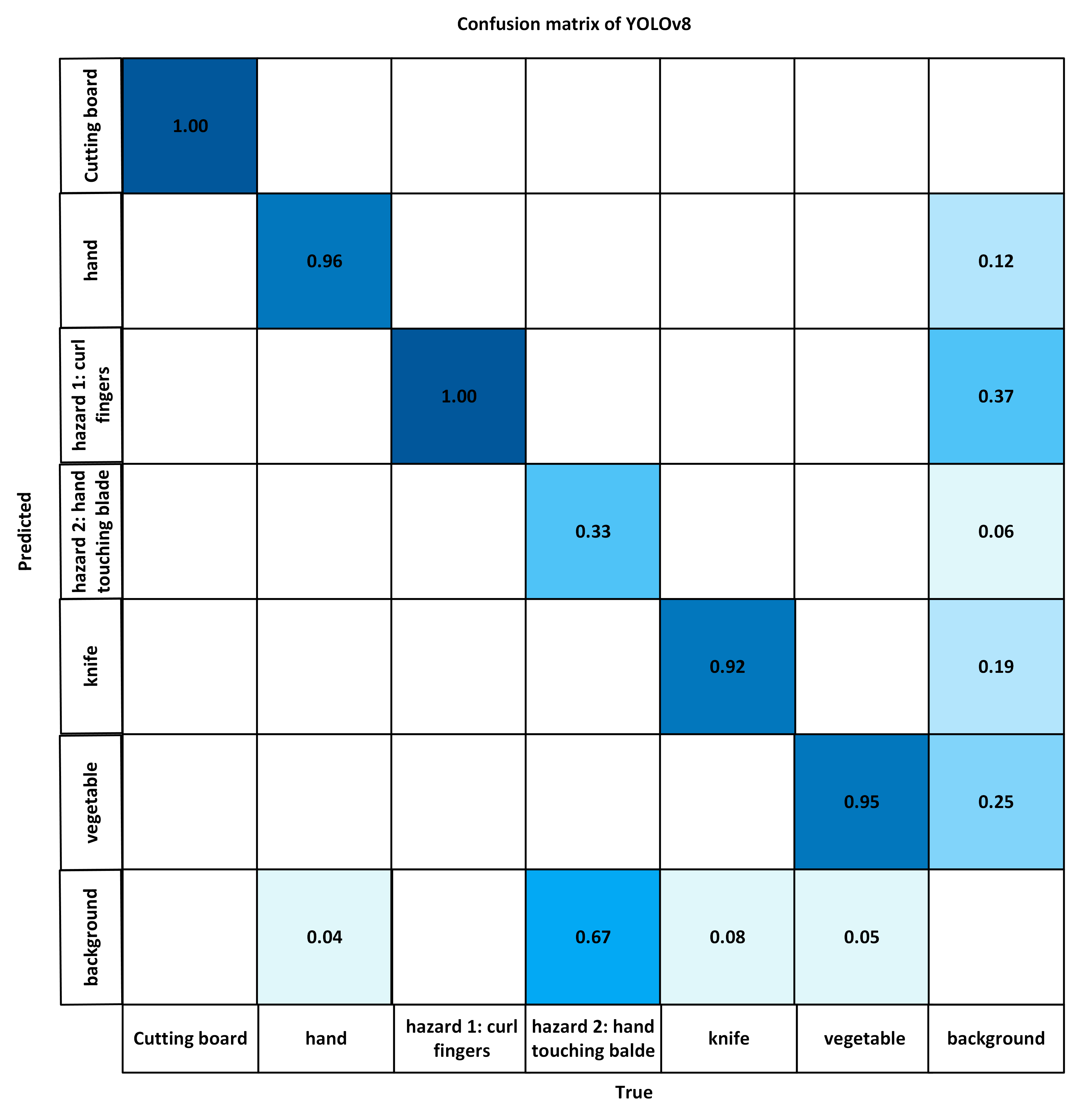}
\end{adjustwidth}
\caption{YOLOv8 normalized confusion matrix.}
\label{Figure:8}
\end{figure}

\begin{figure}[H]
\begin{adjustwidth}{-\extralength}{0cm}
\centering
\includegraphics[height=10cm]{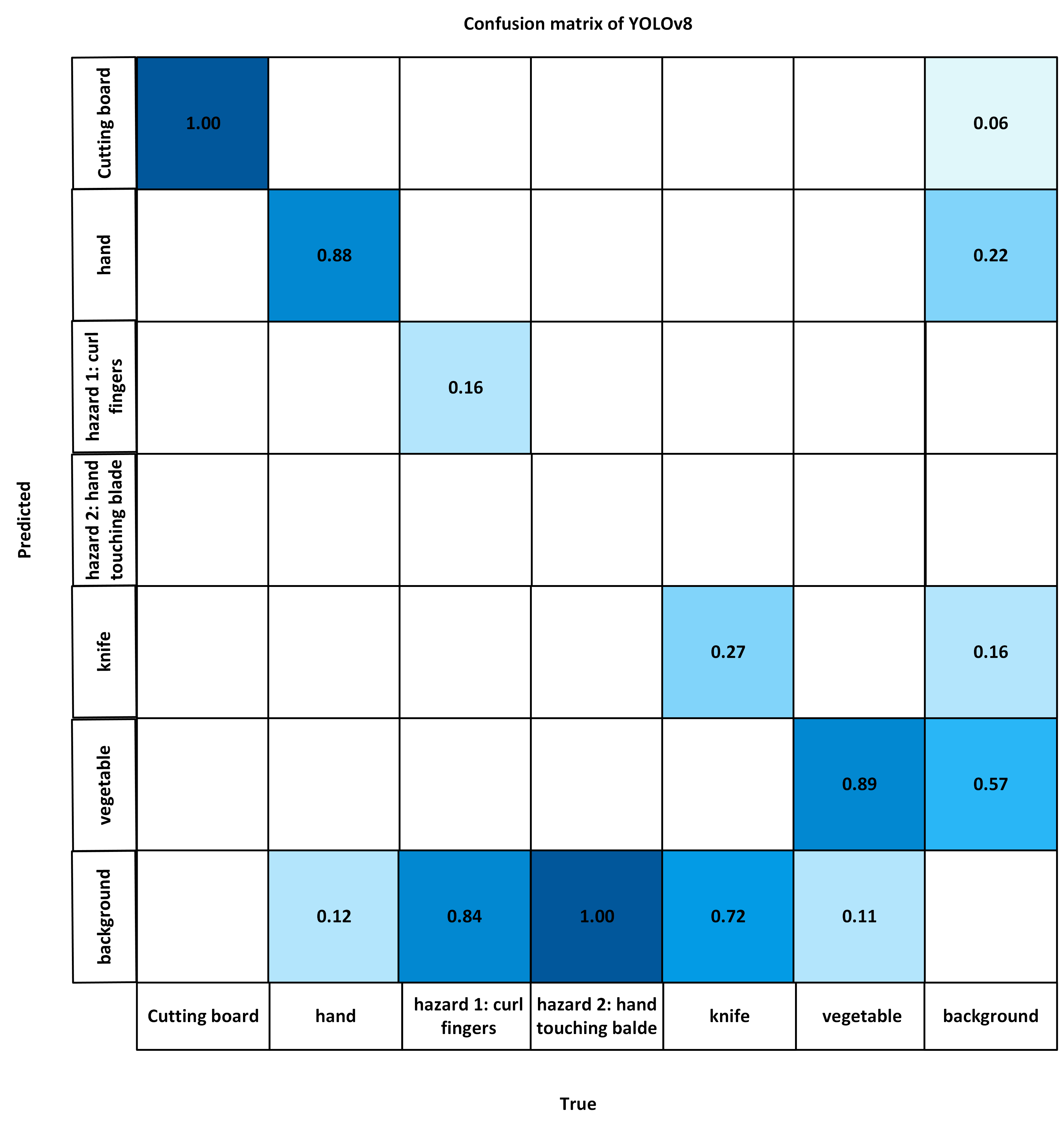}
\end{adjustwidth}
\caption{YOLOv10's normalized confusion matrix.}
\label{Figure:9}
\end{figure}

Finally, background revealed misclassification issues in all models. YOLOv5 had an accuracy of 0.50, whereas YOLOv8  (Figure \ref{Figure:8} and YOLOv10 showed higher misclassification rates with an accuracy of more than 0.50 (0.67 and 1.00, respectively). This clarified that differentiating background elements remained a challenge, with YOLOv10 (Figure \ref{Figure:9} being the worst.

In conclusion, although YOLOv8 had the best performance while classifying hazard 1, it faced challenges while recognizing hazard 2 and background. Also, YOLOv10 had the highest misclassification rates. Hence, YOLOv5 seemed to outperform other models.

\section{Discussion}

This study deals with the comparative analysis of YOLO models, YOLOv5, YOLOv8, and YOLOv10, for safe knife handling in the kitchen. YOLOv8 demonstrated better performance in terms of overall accuracy and recall, achieving high accuracy quickly in the training process. This model also excelled in the classification of hazard 1 with an accuracy of 1.00 compared to YOLOv5 accuracy of 0.50 and YOLOv10 of 0.16. This was due to the YOLOv8 model’s architectural improvement, advanced feature extraction, and more effective handling of object geometries that were capable of identifying the features of hazard 1. Additionally, YOLOv8 added new layers that improved the ability of the model to identify from training data and apply it to real-world scenarios.

However, YOLOv5 showed a balanced performance across various classes, achieving the highest combined F1 score. This model performed well in classifying the cutting board, hand, knife, and vegetable classes with high F1 scores at different confidence levels. YOLOv5’s ability to maintain a balanced accuracy and recall across multiple classes highlighted its reliability in various scenarios. This could be linked to its less complex architecture, detecting the larger objects.

While detecting classes like cutting board and hand, with accuracies of 1.00 and 0.96, respectively, both YOLOv5 and YOLOv8 performed well, denoting that either model could be effective in identifying these classes. These results suggested that YOLOv8 is more flexible for detecting complex objects, in this case, hazard 1, whereas YOLOv5 can be used in applications in which consistent performance in different classes is essential. YOLOv10 does not seem to work well.

While this study provides a detailed comparative analysis of YOLO models (YOLOv5, YOLOv8, and YOLOv10) in knife safety in the kitchen environment, there are several areas that were not explored, such as different kitchen environments along with countertop clutter or different types of food or knives handling. In the future, this study could be further extended to have large datasets with hazards while handling knife.

To summarize, choosing among YOLOv5, YOLOv8, and YOLOv10 involves evaluating their performance characteristics and matching the capability of the model with the requirements of the detection task.

\section{Conclusions}

The performance of YOLOv5, YOLOv8, and YOLOv10 in the environment of knife safety detection has been compared in this study with the pros and cons of all models for five classes. The findings reveal that YOLOv5 outperforms YOLOv8 and YOLOv10 in detecting whether the hand is in contact with the blade, addressing hazard 1. In contrast, YOLOv8 excels in recognizing whether the fingers are curled in the hand holding items, analyzing hazard 2. YOLOv5, YOLOv8, and YOLOv10 demonstrate almost similar accuracy rates in detecting the cutting board. From this study, it is evident that YOLOv5 and YOLOv8 models can detect hazard 2 and hazard 1, respectively, providing significance for the selection of appropriate models on the basis of detection tasks.

Future works can involve increasing the evaluation to assess the performance of the models in the usage of different types of food and knives and working in an environment where objects are partially or fully occluded by utensils and appliances. In addition, the size of the dataset can be increased to include other hazards and classes. This reserach can also be extended to other domains such as renewable energy ~\cite{zahid2023lightweight, hussain2019deployment} and healthcare ~\cite{hussain2023and}.

%%%%%%%%%%%%%%%%%%%%%%%%%%%%%%%%%%%%%%%%%%
\begin{adjustwidth}{-\extralength}{0cm}

\bibliographystyle{unsrt}  % Changes bibliography style to unsorted
\bibliography{ref}  % This points to the filename of your BibTeX file without the .bib extension

\end{adjustwidth}
\end{document}